\documentclass[10pt,conference]{IEEEtran}
\IEEEoverridecommandlockouts 

\usepackage{cite} %
\usepackage{amsmath,amssymb,amsfonts,amsthm}
\usepackage{graphicx}
\usepackage{textcomp}
\usepackage{xcolor}

\usepackage{booktabs} 
\usepackage{dsfont}
\usepackage{algorithm}
\usepackage{algorithmic}
\usepackage{multirow}
\usepackage{wrapfig}
\usepackage{url,enumerate}
\usepackage{mathtools}
\usepackage{xspace}
\usepackage{mathrsfs}
\usepackage{subcaption}
\usepackage{bm}
\usepackage[numbers]{natbib}
\usepackage{hyperref}

\usepackage{enumitem}

\usepackage{tikz}
\usetikzlibrary{positioning, arrows.meta, fit, backgrounds, calc, shapes.geometric}

\newcommand{\ours}{{{COMPOL}}\xspace}

\newcommand{\f}{{\bf f}}
\newcommand{\g}{{\bf g}}

\renewcommand{\v}{{\bf v}}

\newcommand{\y}{{\bf y}}
\newcommand{\z}{{\bf z}}

\newcommand{\Ocal}{\mathcal{O}}

\newcommand{\Fcal}{\mathcal{F}}
\newcommand{\Vcal}{\mathcal{V}}
\newcommand{\Lcal}{\mathcal{L}}

\newcommand{\V}{{\bf V}}

\newcommand{\ben}{\begin{enumerate}}
\newcommand{\een}{\end{enumerate}}

\newcommand{\EE}{\mathbb{E}}

\newcommand{\cmt}[1]{}

\input{formats.sty}

\title{COMPOL: A Scalable Neural Operator Framework for  Multi-Physics Simulations}

\author{
	\IEEEauthorblockN{Junqi Qu\IEEEauthorrefmark{2}, 
		Tao Wang\IEEEauthorrefmark{1}, 
		Yushun Dong\IEEEauthorrefmark{2}, 
		Hewei Tang\IEEEauthorrefmark{1}, 
		and Shibo Li\IEEEauthorrefmark{2}\thanks{*Corresponding author: Shibo Li (shiboli@cs.fsu.edu).}}
	\IEEEauthorblockA{\IEEEauthorrefmark{1}\textit{Hildebrand Dept. of Petroleum and Geosystems Engineering} \\
		\textit{The University of Texas at Austin}, Austin, TX, USA \\
		\{tao.wang, hewei.tang\}@austin.utexas.edu}
	\IEEEauthorblockA{\IEEEauthorrefmark{2}\textit{Department of Computer Science} \\
		\textit{Florida State University}, Tallahassee, FL, USA \\
		\{jq24b, yushun.dong\}@fsu.edu, shiboli@cs.fsu.edu}
}

\begin{document}

\maketitle

\thispagestyle{empty}
\pagestyle{empty}

\begin{abstract}
	Multi-physics simulations play an essential role in accurately modeling complex interactions across diverse scientific and engineering domains. Although neural operators, particularly the Fourier Neural Operator (FNO), have significantly improved computational efficiency, they often fail to capture the intricate correlations inherent in coupled physical processes. To address this limitation, we introduce COMPOL, a novel \underline{co}upled \underline{m}ulti-\underline{p}hysics \underline{o}perator \underline{l}earning framework. COMPOL extends conventional operator architectures by incorporating a sophisticated attention-based aggregation mechanism that effectively models interdependencies among interacting physical processes within latent feature spaces. Our approach is architecture-agnostic and seamlessly integrates into various neural operator frameworks involving latent space transformations. Extensive experiments on diverse benchmarks, including biological reaction-diffusion systems, pattern-forming chemical reactions, multiphase geological flows, and thermo-hydro-mechanical processes, demonstrate that COMPOL consistently achieves superior predictive accuracy compared to state-of-the-art methods. Our code and data are available at \url{https://github.com/AriaQJ/COMPOL}.
\end{abstract}


\section{Introduction}

Physical simulations governed by partial differential equations (PDEs) are fundamental tools across scientific and engineering fields, including aerospace engineering, fluid mechanics, chemical processes, and environmental modeling~\citep{evans2022partial, chorin1990mathematical, bergman2011fundamentals}. These simulations leverage core physical principles, such as conservation laws and symmetries, to predict complex phenomena. Traditional numerical methods, including finite difference, finite element, and finite volume techniques, face significant computational challenges when addressing high-dimensional or coupled multi-physics problems~\citep{quarteroni2010numerical, hughes2003finite}. Recent advances in machine learning have introduced neural operator methods~\citep{kovachki2023neural, lu2021learning, li2020fourier}, which directly approximate solution mappings between infinite-dimensional function spaces, significantly enhancing computational efficiency. Despite their promise, existing neural operator methods often inadequately capture the intricate correlations present in coupled multi-physics systems.

Modeling multi-physics systems presents unique challenges due to the dynamic interactions between multiple distinct physical processes, each governed by its own set of PDEs. These interactions often span multiple spatial and temporal scales, creating significant computational complexity~\citep{keyes2013multiphysics}. Examples include fluid-structure interactions, chemical reaction-diffusion systems, and multiphase geological flows~\citep{bazilevs2013computational, cross2009pattern}. Current neural operator frameworks typically treat these processes independently or apply overly simplistic coupling strategies, resulting in insufficient representation of inter-process dynamics. Consequently, capturing nuanced interactions and accurately predicting system behaviors remains a critical unresolved challenge.

To address these challenges, we propose COMPOL, a novel coupled multi-physics operator learning framework  (see Fig.~\ref{fig:architecture}) explicitly designed to represent complex interactions among multiple physical processes. COMPOL introduces a sophisticated attention-based mechanism~\citep{vaswani2017attention} for latent feature aggregation, capturing detailed inter-process interactions within latent representation spaces. Notably, our framework is architecture-agnostic, capable of integration with various neural operator architectures involving latent feature transformations.

Our contributions are threefold. First, we propose a versatile, architecture-agnostic operator learning framework tailored for coupled multi-physics systems that enhances traditional operator layers with sophisticated feature aggregation. Second, we introduce an attention-based latent aggregation technique that robustly captures dynamic inter-process dependencies across an arbitrary number of physical processes. Third, we demonstrate COMPOL's effectiveness and scalability across diverse multi-physics benchmarks, including predator-prey dynamics, chemical reactions, multiphase flows, and thermo-hydro-mechanical processes, consistently achieving substantial improvements in predictive accuracy compared to state-of-the-art approaches.


\begin{figure*}[t]
	\centering
	\resizebox{0.87\textwidth}{!}{%
		\begin{tikzpicture}[
			node distance=1.2cm and 1.5cm,
			inputNode/.style={font=\large},
			projBlock/.style={rectangle, draw=blue!80, thick, fill=blue!10, minimum width=1.5cm, minimum height=1.2cm, rounded corners, align=center},
			latentNode/.style={font=\large},
			attnBlock/.style={rectangle, draw=orange!80, thick, fill=orange!10, minimum width=3cm, minimum height=1.2cm, rounded corners, align=center},
			opBlock/.style={rectangle, draw=green!60!black, thick, fill=green!10, minimum width=2.5cm, minimum height=1.2cm, rounded corners, align=center},
			arrow/.style={-{Stealth[scale=1.2]}, thick},
			dashedArrow/.style={-{Stealth[scale=1.2]}, thick, dashed, draw=gray!80},
			attnArrow/.style={-{Stealth[scale=1.2]}, thick, draw=orange!90!black}
			]
			
			\node[inputNode] (f1) {$f^1(x)$};
			\node[inputNode, below=1.5cm of f1] (f2) {$f^2(x)$};
			\node[below=0.6cm of f2, font=\Large] (dots1) {$\vdots$};
			\node[inputNode, below=0.6cm of dots1] (fM) {$f^M(x)$};
			
			\node[projBlock, right=1.2cm of f1] (P1) {Projection\\$P^1$};
			\node[projBlock, right=1.2cm of f2] (P2) {Projection\\$P^2$};
			\node[projBlock, right=1.2cm of fM] (PM) {Projection\\$P^M$};
			
			\node[latentNode, right=1.2cm of P1] (v1) {$v_{l-1}^1$};
			\node[latentNode, right=1.2cm of P2] (v2) {$v_{l-1}^2$};
			\node[latentNode, right=1.2cm of PM] (vM) {$v_{l-1}^M$};
			
			\node[opBlock, right=4cm of v1] (h1) {Operator\\$h_l^1$};
			\node[opBlock, right=4cm of v2] (h2) {Operator\\$h_l^2$};
			\node[opBlock, right=4cm of vM] (hM) {Operator\\$h_l^M$};
			
			\path (v2) -- (h2) coordinate[midway] (center);
			\node[attnBlock] at (center) (Attn) {Attention \\ Aggregation $\mathcal{A}$};
			\node[above=0.1cm of Attn, font=\small, color=orange!90!black] (z_label) {Global Interaction $z_{l-1}$};
			
			\node[latentNode, right=1.5cm of h1] (v1_next) {$v_l^1$};
			\node[latentNode, right=1.5cm of h2] (v2_next) {$v_l^2$};
			\node[latentNode, right=1.5cm of hM] (vM_next) {$v_l^M$};
			
			\node[projBlock, right=2cm of v1_next] (Q1) {Projection\\$Q^1$};
			\node[projBlock, right=2cm of v2_next] (Q2) {Projection\\$Q^2$};
			\node[projBlock, right=2cm of vM_next] (QM) {Projection\\$Q^M$};
			
			\node[inputNode, right=1.2cm of Q1] (g1) {$g^1(x)$};
			\node[inputNode, right=1.2cm of Q2] (g2) {$g^2(x)$};
			\node[below=0.6cm of g2, font=\Large] (dots2) {$\vdots$};
			\node[inputNode, right=1.2cm of QM] (gM) {$g^M(x)$};

			
			\draw[arrow] (f1) -- (P1);
			\draw[arrow] (f2) -- (P2);
			\draw[arrow] (fM) -- (PM);
			
			\draw[arrow] (P1) -- (v1);
			\draw[arrow] (P2) -- (v2);
			\draw[arrow] (PM) -- (vM);
			
			\draw[arrow] (v1) -- (h1);
			\draw[arrow] (v2) -- (h2);
			\draw[arrow] (vM) -- (hM);
			
			\draw[dashedArrow] (v1) -- (Attn.north west);
			\draw[dashedArrow] (v2) -- (Attn.west);
			\draw[dashedArrow] (vM) -- (Attn.south west);
			
			\draw[attnArrow] (Attn.north east) -- (h1.south west);
			\draw[attnArrow] (Attn.east) -- (h2.west);
			\draw[attnArrow] (Attn.south east) -- (hM.north west);
			
			\draw[arrow] (h1) -- (v1_next);
			\draw[arrow] (h2) -- (v2_next);
			\draw[arrow] (hM) -- (vM_next);
			
			\draw[arrow, dashed] (v1_next) -- (Q1) node[midway, above, font=\small] {$L$ Layers};
			\draw[arrow, dashed] (v2_next) -- (Q2);
			\draw[arrow, dashed] (vM_next) -- (QM);
			
			\draw[arrow] (Q1) -- (g1);
			\draw[arrow] (Q2) -- (g2);
			\draw[arrow] (QM) -- (gM);
			
			\begin{scope}[on background layer]
				\node[
				rectangle, 
				draw=black!50, 
				dashed, 
				thick, 
				rounded corners=15pt, 
				fill=black!3, 
				inner sep=15pt, 
				fit=(v1) (hM) (Attn) (vM_next) (z_label)
				] (layerBox) {};
				\node[anchor=south east, font=\bfseries\large, text=black!70] at (layerBox.south east) {COMPOL Layer $l$};
			\end{scope}
			
		\end{tikzpicture}
	}
	\caption{\small Architecture of the Coupled Multi-Physics Operator Learning (COMPOL) framework. The model processes $M$ interacting physical systems in parallel. Initial physical input functions $f^m(x)$ are mapped into a latent space via process-specific linear projections $P^m$. Within each of the $L$ COMPOL layers, the latent representations $v_{l-1}^m$ undergo transformations via independent neural operators $h_l^m$ (e.g., FNO layers). Simultaneously, an attention-based aggregation mechanism $\mathcal{A}$ evaluates the intermediate states to compute a global interaction feature $z_{l-1}$. This aggregated state explicitly captures the nonlinear cross-process dependencies and is fed into the respective neural operators to produce the updated latent states $v_l^m$. After $L$ iterative layers, dedicated linear projections $Q^m$ reconstruct the latent representations back into the final physical solution spaces $g^m(x)$.}
	\label{fig:architecture}
\end{figure*}

\section{Background and Related Work}\label{sect:bk}

\paragraph{Operator Learning}
Operator learning~\cite{kovachki2023neural} directly approximates solution operators of partial differential equations (PDEs) by learning mappings between infinite-dimensional function spaces. Consider a PDE defined as: $\mathfrak{L}(u)(x) = f(x), x\in\Omega; u(x)=0, x\in\partial\Omega$
where $\mathfrak{L}$ is a differential operator, $u$ is the solution function, $f$ is the input function, $\Omega$ is the problem domain and $\partial\Omega$ represents its boundary. Operator learning aims to approximate the inverse operator $\mathfrak{L}^{-1}$ mapping inputs $f$ to $u$ through a parametric operator $\psi_\theta: \mathcal{H} \rightarrow \mathcal{Y}$, minimizing the empirical risk over function pairs $\{(f_n, u_n)\}_{n=1}^N$: $\theta^*=\operatorname{argmin}_\theta \frac{1}{N} \sum_{n=1}^N\|\psi_\theta(f_n)-u_n\|_{\mathcal{Y}}^2$
where $\mathcal{H}$ and $\mathcal{Y}$ represent suitable function spaces, such as Banach or Hilbert spaces.

\paragraph{Neural Operator Architectures}
Several neural operator architectures have been proposed. The Fourier Neural Operator (FNO)~\citep{li2020fourier} approximates solution operators using spectral convolutions: $v(x) \leftarrow \sigma\left(W v(x)+\mathcal{F}^{-1}(R(k) \cdot \mathcal{F}(v(x)))\right),$
where $\Fcal$ and $\Fcal^{-1}$ denote the Fourier transform and its inverse, and $R(k)$ is a learned spectral kernel. The U-shaped FNO (UFNO)~\citep{WEN2022104180} extends FNO with a U-Net-style encoder-decoder for improved multi-scale feature extraction. The Graph Neural Operator (GNO)~\citep{li2020neural} combines Nystr\"{o}m approximation with graph neural networks for function convolution. The Deep Operator Net (DeepONet)~\citep{lu2021learning} employs a dual-network architecture with branch and trunk networks, later refined into POD-DeepONet~\citep{lu2022comprehensive} using PCA bases for improved stability. Transolver~\citep{wu2024transolver} introduces a geometry-aware transformer-based operator for PDEs on general meshes. Recent advances have also developed mesh-agnostic approaches~\citep{yin2022continuous, boussif2022magnet} that reformulate PDE simulation as ODE-solving problems. Training neural operators often requires extensive high-quality data, motivating multi-fidelity modeling~\citep{tang2024multi} and active learning approaches~\citep{pmlr-v151-li22b, li2023multiresolution}.

\paragraph{Multi-Physics Neural Operators}
Operator learning extends traditional surrogate modeling that focused on mapping system parameters to solutions~\citep{higdon2008computer, zhe2019scalable}. Recent work has explored neural operators for multi-physics settings~\citep{mccabe2023multiple, rahman2024pretraining, hao2024dpot}, though these target general physics pre-training rather than interactions within individual coupled systems. Closely related is the Coupled Multiwavelet Neural Operator (CMWNO)~\citep{xiao2023coupled}, which models coupled PDEs using wavelet-based methods but has limitations including rigid decomposition schemes and difficulty scaling to multiple interacting processes or higher-dimensional domains. Our method builds upon flexible FNO layers, and the introduced attention-based aggregation mechanism integrates seamlessly with any latent-feature neural operator framework while naturally supporting complex interactions across multiple physical processes.

\section{Method}

\subsection{Multi-Physics Simulation}

We consider multi-physics systems governed by coupled partial differential equations (PDEs), which naturally emerge in diverse scientific and engineering disciplines. These systems involve multiple interacting physical processes, each described by its own PDE and interconnected through nonlinear coupling terms~\citep{weinan2003multiscale}. Such interactions substantially influence overall system behavior, creating considerable modeling and computational challenges. Formally, a coupled PDE system involving $M$ interacting physical processes is given by:
\begin{equation}
	\begin{aligned}
		\mathfrak{L}_m\left(u_1, u_2, \ldots, u_M\right)(x, t) & =f_m(x, t), \quad m=1, \ldots, M, \\
		u_m(x, t) & =0, \quad x \in \partial \Omega_m,
	\end{aligned}
\end{equation}
where $\mathfrak{L}_m$ represents differential operators, potentially including spatial and temporal derivatives and nonlinear terms capturing interactions among physical processes. Here, $u_m(x,t)$ are the unknown state variables associated with the $m$-th physical process, and $f_m(x,t)$ represent known source terms, external inputs, or boundary conditions.

\subsection{Coupled Multi-Physics Operator Learning (\ours)}

Neural operator learning has shown great promise for modeling physical phenomena governed by PDEs. However, most existing approaches assume systems are governed by a single set of PDEs, overlooking the reality where multiple processes interact across different scales~\citep{keyes2013multiphysics}. Existing techniques like feature concatenation~\citep{WEN2022104180} and cross-overlaying~\citep{xiao2023coupled} attempt to model these interactions but are often insufficient for sophisticated multi-physics scenarios.

To overcome this limitation, we propose \ours, a coupled multi-physics neural operator $\mathcal{MPO}(\Ocal_1, \cdots, \Ocal_M)$ designed to capture intricate correlations among individual operators $\Ocal_1, \cdots, \Ocal_M$. COMPOL approximates solution operators for coupled PDE systems as mappings between function spaces: $\mathcal{G}_\theta: \mathcal{H}_1 \times \mathcal{H}_2 \times \cdots \times \mathcal{H}_M \rightarrow \mathcal{Y}_1 \times \mathcal{Y}_2 \times \cdots \times \mathcal{Y}_M$, where each $\mathcal{H}_m$ denotes the input function space and $\mathcal{Y}_m$ denotes the output solution function space for the $m$-th physical process. Given input functions $f^m(x)$ representing initial conditions or external inputs, we first embed each process into a shared latent space: $v_0^m(x)=P^m(f^m(x))$, where $P^m$ are linear projection operators mapping inputs into a common latent feature space.

Subsequently, latent features undergo iterative transformations using nonlinear multi-physics neural operator layers: $v_l^m(x)=h_l^m\left(v_{l-1}^m(x), z_{l-1}(x)\right), \quad l=1, \ldots, L$, where each layer $h_l^m$ comprises a nonlinear transformation representing internal process dynamics, augmented by interaction-aware aggregated features $z_{l-1}(x)$. The latent representation $v_l^m(x)$ captures process-specific evolution and local dynamics independently for each physical process, while the aggregated feature $z_{l-1}(x)$ explicitly represents inter-process interactions, integrating global context across different processes. This separation is a crucial design choice, allowing the model to independently refine representations of each process while explicitly modeling nonlinear interactions among them.

The aggregated interaction features are computed using an attention-based aggregation function: $z_{l-1}(x)=\mathcal{A}(v_{l-1}^1(x), v_{l-1}^2(x), \ldots, v_{l-1}^M(x))$, ensuring coherent representation of inter-process dependencies. Finally, each process-specific solution is reconstructed via dedicated linear projection operators: $g^m(x)=Q^m\left(v_L^m(x)\right)$, yielding predicted solutions in their respective physical spaces.

\subsection{Aggregation of Multi-Physics Operator Layers}

We now present the formulation using discretized function notation. Consider a coupled physical system with $M$ processes, each associated with discretized input functions $\f^1, \dots, \f^M$. Each process $m$ maps its input $\f^m$ to a latent representation $\v_0^m$ via a channel-wise linear layer $P^m: \mathbb{R}^{d_m}\rightarrow\mathbb{R}^{d_h}$, where $d_m$ and $d_h$ are input and latent dimensions, respectively. By stacking $L$ neural operator layers, each layer $h_l^m$ transforms latent representations $\v_{l-1}^m$ into $\v_l^m$. A final linear layer $Q^m: \mathbb{R}^{d_h}\rightarrow\mathbb{R}^{d_m}$ maps the latent representation $\v_L^m$ back to the output function $\g^m$. To capture inter-process interactions, we compute an aggregated latent state $\z_l=\mathcal{A}(\{\{\v^m_j\}_{m=1}^M\}_{j=0}^l)$ at each layer using an attention-based mechanism~\citep{vaswani2017attention}.

For notational convenience, let $\Vcal = \{\{\v^m_l\}_{m=1}^M\}_{l=0}^L$ represent the complete collection of intermediate latent features across $M$ processes and $L+1$ layers. We define $\V_l$ as the $M \times d_h$ matrix obtained by extracting the $l$-th slice along the first dimension of $\Vcal$, and $\Vcal_{l} = \{\V_j\}_{j=0}^{l}$ represents the sequence of latent functions from layer $0$ through layer $l$.

\paragraph{Attention-Based Aggregation} We compute the global latent interaction state through scaled dot-product attention tailored for multi-physics systems. The latent features from each process are linearly projected into query ($Q$), key ($K$), and value ($A$) representations via process-specific transformations: $Q=\Phi_Q(\Vcal_l)$, $K=\Phi_K(\Vcal_l)$, and $A=\Phi_A(\Vcal_l)$. The aggregation state $\z_l$ is derived as a weighted sum of value vectors, where each weight $\alpha_j$ quantifies the significance of interactions involving the $j$-th process:
\begin{equation}
	z_l=\sum_{j=1}^M \alpha_j A_j, \quad \alpha_j=\frac{\exp \left(Q K_j^{\top} / \sqrt{d_k}\right)}{\sum_{i=1}^M \exp \left(Q K_i^{\top} / \sqrt{d_k}\right)}
\end{equation}
with $d_k$ representing the dimensionality of key vectors. This approach explicitly encodes process-specific interactions, dynamically identifying the most influential cross-process features at each layer.

\paragraph{Complexity Analysis} The computational complexity of COMPOL includes three components. Projection layers have complexity $O(M d_m d_h)$, with $M$ processes, input dimension $d_m$, and latent dimension $d_h$. Each FNO layer involves FFT operations with complexity $O(N \log N)$ per channel~\citep{cooley1965algorithm}, where $N$ denotes spatial discretization points, yielding total complexity $O(L M d_h \cdot N \log N)$ for $L$ layers. The attention aggregation contributes $O(L M^2 d_h)$. Overall complexity is dominated by Fourier operations in typical scenarios where $N \log N \gg M, d_h$.

\paragraph{Generalizability to Other Neural Operator Frameworks} \ours builds upon FNO due to its proven efficiency, but the proposed aggregation mechanism can be seamlessly integrated with other frameworks, including DeepONet~\citep{lu2021learning}, GNO~\citep{li2020neural}, and transformer-based operators~\citep{wu2024transolver}.

\subsection{Training}

Given training data from a coupled multi-physics system, denoted as $\{\{\f_n^m, \y_n^m\}_{n=1}^{N_m}\}_{m=1}^M$, where $\f_n^m$ and $\y_n^m$ represent the $n$-th input and output functions from the $m$-th process, we optimize our coupled multi-physics neural operator by minimizing the empirical risk:
$
	\Lcal_{\mathcal{MPO}} = \EE_{m\sim\pi}\EE_{f^m\sim\mu^m}||\mathcal{MPO}(f)-y^m|| \approx \frac{1}{M}\sum_{m=1}^M\frac{1}{N_m}\sum_{n=1}^{N_m}||\g^m_n-\y_n^m||
$
where $\g_n^m$ is the prediction of $\f^m_n$. We use gradient-based optimization to minimize $\Lcal_{\mathcal{MPO}}$.

\begin{table*}[!htbp]
	\centering
	\renewcommand{\arraystretch}{1.15}
	\setlength{\tabcolsep}{4pt}
	\resizebox{0.82\linewidth}{!}{%
		\begin{tabular}{@{}ll ccccc@{}}
			\toprule
			\textbf{Method} & $N_{\text{train}}$ & \textbf{Lotka-Volterra} & \textbf{Belousov-Zhab.} & \textbf{Gray-Scott} & \textbf{Multiphase} & \textbf{THM} \\
			\midrule
			\multirow{2}{*}{$\text{FNO}_{c}$}
			& 512  & 0.3251{\scriptsize$\pm$0.0507} & 0.0058{\scriptsize$\pm$1.4e-4} & 0.0056{\scriptsize$\pm$1.0e-4} & 0.0232{\scriptsize$\pm$5.0e-4} & 0.1826{\scriptsize$\pm$0.0088} \\
			& 1024 & 0.1499{\scriptsize$\pm$0.0097} & 0.0022{\scriptsize$\pm$4.4e-5} & 0.0042{\scriptsize$\pm$7.8e-5} & 0.0684{\scriptsize$\pm$0.1071} & 0.1577{\scriptsize$\pm$0.0073} \\
			\midrule
			\multirow{2}{*}{CMWNO}
			& 512  & 0.3989{\scriptsize$\pm$0.0422} & 0.0375{\scriptsize$\pm$0.0015} & \multicolumn{1}{c}{--} & \multicolumn{1}{c}{--} & \multicolumn{1}{c}{--} \\
			& 1024 & 0.3876{\scriptsize$\pm$0.0357} & 0.0336{\scriptsize$\pm$0.0019} & \multicolumn{1}{c}{--} & \multicolumn{1}{c}{--} & \multicolumn{1}{c}{--} \\
			\midrule
			\multirow{2}{*}{$\text{UFNO}_{c}$}
			& 512  & 0.2519{\scriptsize$\pm$0.0518} & 0.0239{\scriptsize$\pm$7.7e-4} & 0.0083{\scriptsize$\pm$3.0e-4} & 0.0241{\scriptsize$\pm$4.0e-4} & 0.1824{\scriptsize$\pm$0.0094} \\
			& 1024 & 0.1677{\scriptsize$\pm$0.0120} & 0.0122{\scriptsize$\pm$5.1e-4} & 0.0044{\scriptsize$\pm$9.8e-5} & 0.0137{\scriptsize$\pm$0.0010} & 0.1496{\scriptsize$\pm$0.0086} \\
			\midrule
			\multirow{2}{*}{$\text{Transolver}_{c}$}
			& 512  & 0.2606{\scriptsize$\pm$0.0515} & 0.0329{\scriptsize$\pm$0.0089} & 0.0251{\scriptsize$\pm$9.0e-4} & 0.0503{\scriptsize$\pm$0.0311} & 0.2176{\scriptsize$\pm$0.0123} \\
			& 1024 & 0.1262{\scriptsize$\pm$0.0348} & 0.0151{\scriptsize$\pm$0.0023} & 0.0174{\scriptsize$\pm$9.0e-4} & 0.0185{\scriptsize$\pm$0.0037} & 0.1916{\scriptsize$\pm$0.0126} \\
			\midrule
			\multirow{2}{*}{$\text{DeepONet}_{c}$}
			& 512  & 0.4024{\scriptsize$\pm$0.0143} & 0.6491{\scriptsize$\pm$0.0199} & 0.7655{\scriptsize$\pm$0.0039} & 0.1123{\scriptsize$\pm$0.0051} & 0.2867{\scriptsize$\pm$0.0199} \\
			& 1024 & 0.3005{\scriptsize$\pm$0.0285} & 0.5810{\scriptsize$\pm$0.0352} & 0.6700{\scriptsize$\pm$0.0240} & 0.0766{\scriptsize$\pm$0.0049} & 0.2388{\scriptsize$\pm$0.0125} \\
			\midrule
			\multirow{2}{*}{\textbf{\ours}}
			& 512  & \textbf{0.0918}{\scriptsize$\pm$0.0042} & \textbf{0.0016}{\scriptsize$\pm$2.7e-5} & \textbf{0.0050}{\scriptsize$\pm$0.0016} & \textbf{0.0182}{\scriptsize$\pm$0.0011} & \textbf{0.1647}{\scriptsize$\pm$0.0092} \\
			& 1024 & \textbf{0.0510}{\scriptsize$\pm$0.0048} & \textbf{0.0008}{\scriptsize$\pm$2.6e-5} & \textbf{0.0035}{\scriptsize$\pm$7.2e-5} & \textbf{0.0110}{\scriptsize$\pm$9.0e-4} & \textbf{0.1442}{\scriptsize$\pm$0.0066} \\
			\bottomrule
	\end{tabular}}
	\caption{ Comparison of relative $L_2$ prediction errors (mean $\pm$ standard deviation over 5 runs), defined as $\frac{\|\hat{u} - u\|_2}{\|u\|_2}$, where $\hat{u}$ and $u$ denote the predicted and ground-truth solutions, respectively for COMPOL against baseline neural operators (FNO, UFNO, Transolver, DeepONet, CMWNO) across five multi-physics benchmark datasets, evaluated with 512 and 1024 training samples. Bold values indicate the best performance for each case.}
\label{tab:nrmse}
\end{table*}

\begin{figure}[!htbp]
	\centering
	\setlength\tabcolsep{0.01pt}
	\captionsetup[subfigure]{aboveskip=0pt,belowskip=0pt}
	\begin{tabular}[c]{c}
		\begin{subfigure}[t]{0.47\textwidth}
			\centering
			\includegraphics[width=\textwidth]{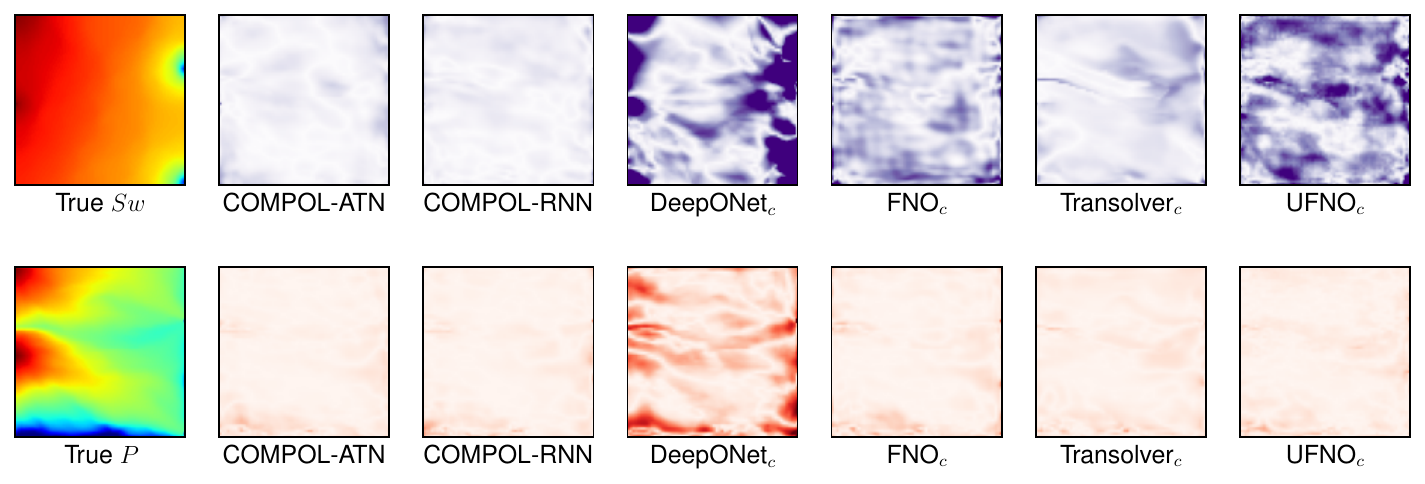}
		\end{subfigure} 
		\\
		\begin{subfigure}[t]{0.47\textwidth}
			\centering
			\includegraphics[width=\textwidth]{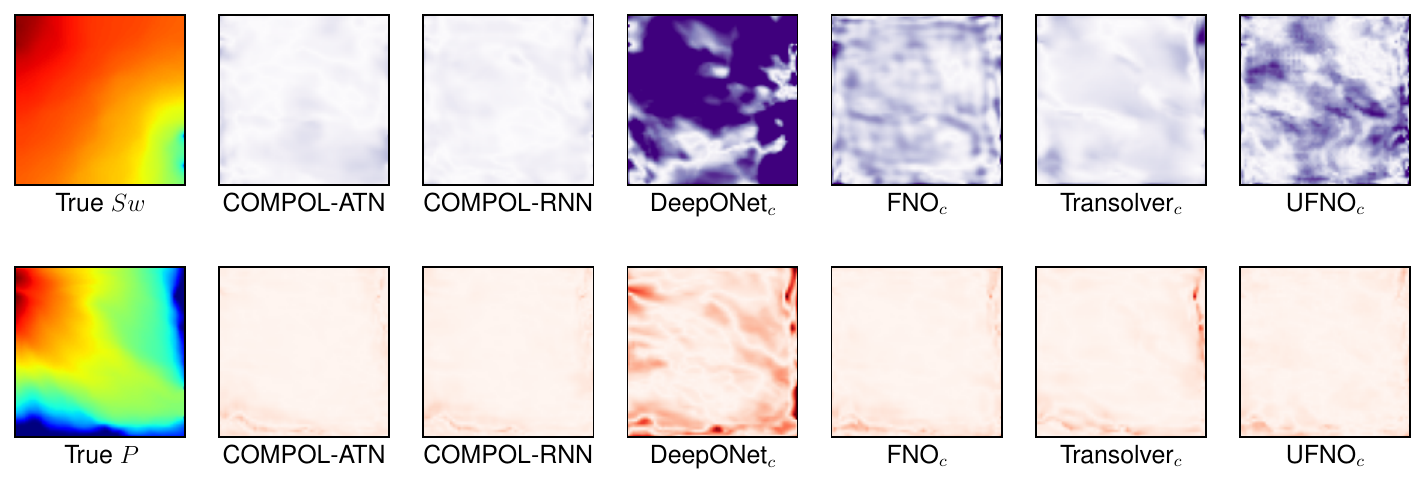}
		\end{subfigure} 
	\end{tabular}
	\caption{\small Element-wise solution error of coupled multiphase flow. The left most column is the ground-truth phase pressure $P$ and saturation $S_w$. The other columns are the absolute errors of each method. The lighter the color, the smaller the error.} 	
	\label{fig:vis_errs_mf}
\end{figure}

\section{Experiment}

\paragraph{Benchmarks} We evaluated our framework on five coupled multi-physics PDE systems spanning diverse scientific applications. (1) The \textit{1-D Lotka-Volterra equations}~\citep{murray2007mathematical} model predator-prey dynamics through two coupled reaction-diffusion processes with diffusion coefficients $D_u = D_v = 0.01$ and interaction parameters $a = b = c = d = 0.01$, initialized with Gaussian random fields (length-scale $l=0.1$) under periodic boundary conditions. (2) The \textit{1-D Belousov-Zhabotinsky equations}~\citep{petrov1993controlling} describe chemical oscillations with three reactants $(u, v, w)$ governed by nonlinear coupling terms including $uv$ and $u^2$, using diffusion coefficients $\epsilon_1 = \epsilon_2 = 10^{-2}$ and $\epsilon_3 = 5 \times 10^{-3}$, simulated from $t=0$ to $t=0.5$. (3) The \textit{2-D Gray-Scott equations}~\citep{pearson1993complex} model pattern formation in chemical reactions with activator-inhibitor dynamics, using $D_u = 0.12$, $D_v = 0.06$, feeding rate $F = 0.054$, and removal rate $k = 0.063$, evolved to $T = 20$. (4) The \textit{2-D multiphase flow problem}~\citep{bear2010modeling} simulates oil-water two-phase flow in porous media with permeability ranging from 1mD to 1000mD sampled from fractal distributions, tracking phase pressure and saturation over $7.5 \times 10^6$ seconds using GEOS simulator. (5) The \textit{2-D Thermo-Hydro-Mechanical (THM) problem}~\citep{gao2020three} couples five processes (pore pressure, temperature, and three strain components) governed by momentum balance, fluid mass balance, and heat transfer equations, with permeability fields generated using fractal algorithms ($k \in [10, 200]$ D). Training datasets were generated using numerical solvers with 256-point meshes for 1-D cases and $64 \times 64$ meshes for 2-D cases. We trained models on two dataset sizes (512 and 1024 samples) and evaluated on 200 independent test examples. Coefficients and boundary conditions remained constant to isolate effects of varying initial conditions. The primary objective was to assess our framework's capability to map initial conditions ($t=0$) to final solution fields ($t=T$), demonstrating its ability to model complex multi-physics interactions. 

\begin{table*}[!htbp]
	\centering
	
	\renewcommand{\arraystretch}{1.15}
	\setlength{\tabcolsep}{4pt}
	\resizebox{0.7\linewidth}{!}{%
		\begin{tabular}{@{}lcccc cc@{}}
			\toprule
			\textbf{Method} & \textbf{Modes} & \textbf{Width} & \textbf{Layers} & \textbf{Params (M)} & \textbf{BZ} & \textbf{LV} \\
			\midrule
			$\text{FNO}_{c}$ (default) & 16 & 64 & 4 & 0.58 & 0.0022{\scriptsize$\pm$4.4e-5} & 0.1499{\scriptsize$\pm$0.0097} \\
			$\text{FNO}_{c}$ & 96 & 64 & 4 & 3.20 & 0.0018{\scriptsize$\pm$4.9e-5} & 0.1183{\scriptsize$\pm$0.0106} \\
			$\text{FNO}_{c}$ & 64 & 64 & 6 & 3.23 & 0.0020{\scriptsize$\pm$4.4e-5} & 0.1303{\scriptsize$\pm$0.0152} \\
			$\text{FNO}_{c}$ & 48 & 64 & 8 & 3.25 & 0.0032{\scriptsize$\pm$1.0e-4} & 0.1488{\scriptsize$\pm$0.0160} \\
			$\text{FNO}_{c}$ & 16 & 224 & 2 & 3.62 & 0.0014{\scriptsize$\pm$2.6e-5} & 0.1757{\scriptsize$\pm$0.0102} \\
			$\text{FNO}_{c}$ & 16 & 128 & 6 & 3.48 & 0.0018{\scriptsize$\pm$3.8e-5} & 0.1761{\scriptsize$\pm$0.0099} \\
			$\text{FNO}_{c}$ & 16 & 160 & 4 & 3.64 & 0.0016{\scriptsize$\pm$4.0e-5} & 0.1698{\scriptsize$\pm$0.0090} \\
			\midrule
			\textbf{\ours} & 16 & 64 & 4 & 2.10\,/\,1.49 & \textbf{0.0008}{\scriptsize$\pm$2.6e-5} & \textbf{0.0510}{\scriptsize$\pm$0.0048} \\
			\bottomrule
	\end{tabular}}
	\caption{Comparison of relative $L_2$ errors for parameter-matched FNO variants versus COMPOL on Belousov–Zhabotinsky (BZ) and Lotka–Volterra (LV) with 1024 training samples. COMPOL's parameter counts are reported as BZ / LV. Despite having fewer parameters, COMPOL outperforms all FNO variants, demonstrating that its advantage stems from explicit cross-process coupling rather than increased capacity.}
	\label{tab:ablation_params}
\end{table*}

\begin{table}[!htbp]
	\centering
	
	\renewcommand{\arraystretch}{1.15}
	\setlength{\tabcolsep}{6pt}
	\resizebox{0.77\linewidth}{!}{%
		\begin{tabular}{@{}l cc@{}}
			\toprule
			\textbf{Method} & $\boldsymbol{N_{\text{train}}=512}$ & $\boldsymbol{N_{\text{train}}=1024}$ \\
			\midrule
			FNO        & 0.0715{\scriptsize$\pm$0.0033} & 0.0422{\scriptsize$\pm$0.0037} \\
			CMWNO      & 0.4441{\scriptsize$\pm$0.0134} & 0.4509{\scriptsize$\pm$0.0168} \\
			UFNO       & 0.1043{\scriptsize$\pm$0.0061} & 0.0565{\scriptsize$\pm$0.0012} \\
			Transolver & 0.1264{\scriptsize$\pm$0.0733} & 0.4066{\scriptsize$\pm$0.4248} \\
			DeepONet   & 0.4007{\scriptsize$\pm$0.0216} & 0.3749{\scriptsize$\pm$0.0181} \\
			\midrule
			\textbf{\ours} & \textbf{0.0200}{\scriptsize$\pm$0.0021} & \textbf{0.0106}{\scriptsize$\pm$0.0006} \\
			\bottomrule
		\end{tabular}
	}
	\caption{Comparison of prediction accuracy (relative $L_2$ error) for the single-process Burgers' equation trained with 512 and 1024 examples. Results averaged over 5 runs, demonstrating that COMPOL's attention-based aggregation improves performance even in single-process scenarios.}
	\label{tab:ablation_single}
\end{table}

\begin{table}[!htbp]
	\centering
	\renewcommand{\arraystretch}{1.15}
	\setlength{\tabcolsep}{5pt}
	\resizebox{0.86\linewidth}{!}{%
		\begin{tabular}{@{}l ccc cc@{}}
			\toprule
			\multirow{2}{*}{\textbf{Method}}
			& \multicolumn{3}{c}{\textbf{BZ} $(u, v, w)$}
			& \multicolumn{2}{c}{\textbf{GS} $(u, v)$} \\
			\cmidrule(lr){2-4}\cmidrule(l){5-6}
			& $u$ & $v$ & $w$ & $u$ & $v$ \\
			\midrule
			CMWNO       & 0.6517 & 0.7629 & 0.5010 & --     & --     \\
			DeepONet    & 2.0934 & 3.4638 & 2.7937 & 31.14  & 30.57  \\
			FNO         & 0.1475 & 0.0930 & 0.1182 & 0.4024 & 0.2256 \\
			Transolver  & 0.3899 & 0.4728 & 0.2775 & 0.4859 & 0.3413 \\
			UFNO        & 0.2210 & 0.1786 & \textbf{0.1072} & 0.4056 & 0.2734 \\
			\midrule
			\textbf{\ours} & \textbf{0.1403} & \textbf{0.0672} & 0.1091 & \textbf{0.4037} & \textbf{0.2147} \\
			\bottomrule
		\end{tabular}
	}
	\caption{$\text{MAE} = \frac{1}{K}\sum_{k=1}^{K}|E_{\hat{u}}(k) - E_u(k)|$ of spectral energy distributions for Belousov–Zhabotinsky (BZ) and Gray–Scott (GS) systems. Lower values indicate better agreement with ground-truth spectra; best per column in \textbf{bold}.}
	\label{tab:spectral-energy-mae}
\end{table}

\paragraph{Competing Methods} We compare COMPOL against state-of-the-art neural operators: FNO~\citep{li2020fourier}, UFNO~\citep{WEN2022104180}, Transolver~\citep{wu2024transolver}, DeepONet~\citep{lu2021learning}, and CMWNO~\citep{xiao2023coupled}, each using their official implementations. Except for CMWNO, these baselines are not inherently designed for multi-physics modeling; we adapt them by concatenating process inputs and outputs across channels (denoted with subscript `c', e.g., FNO\textsubscript{c}). All models were implemented in PyTorch~\citep{paszke2019pytorch} and trained with the Adam optimizer (learning rate $= 0.001$). CMWNO employed a step scheduler following its original implementation, while all other models used cosine annealing~\citep{loshchilov2016sgdr}. Models were trained for 500 epochs, sufficient for convergence except DeepONet which required 10,000 epochs. Experiments were conducted on NVIDIA A100 GPUs with 5-fold cross-validation, and performance was evaluated using average relative $L_2$ error with standard deviations. CMWNO's wavelet-based coupling relies on a fixed multiwavelet decomposition scheme that is implemented only for 1-D domains in its released codebase; extending it to 2-D grids is non-trivial and beyond the scope of this work. We therefore report CMWNO results only on the two 1-D benchmarks.

\subsection{Analysis of Predictive Performance}

The experimental results in Table~\ref{tab:nrmse} present the predictive performance of COMPOL compared to baseline neural operators (FNO, UFNO, Transolver, DeepONet, and CMWNO), evaluated for both 512 and 1024 training samples. For 512 training samples, COMPOL consistently achieves the lowest prediction errors, outperforming baseline methods with improvements up to 72.41\%. When increasing the training set size to 1024 samples, COMPOL maintains superior performance, achieving improvements up to 63.64\% compared to baselines, notably excelling on the Lotka-Volterra and Belousov-Zhabotinsky datasets. Baseline neural operators adapted for multi-physics tasks exhibit significantly higher errors at both training scales. CMWNO, despite being explicitly designed for multi-physics modeling, consistently underperforms relative to COMPOL across all benchmarks. These results affirm the superior effectiveness and scalability of COMPOL in multi-physics modeling scenarios. Fig.~\ref{fig:vis_errs_mf} visualizes the element-wise absolute errors for the multiphase flow benchmark. COMPOL produces substantially lower errors across both phase pressure and saturation fields, while baselines such as DeepONet exhibit large localized errors.

\subsection{Ablation Study of Coupling Mechanisms}

\paragraph{Capacity vs. Coupling: Parameter-Matched FNO}
COMPOL instantiates a separate neural operator per physical process and aggregates cross-process information in latent space; consequently, its parameter count scales roughly linearly with the number of processes $M$ relative to a single FNO with naïve channel concatenation. To test whether our gains are merely due to increased capacity, we scale FNO to match COMPOL's parameter count by increasing Fourier modes, network width, and depth. On Belousov–Zhabotinsky (BZ; 3 processes) and Lotka–Volterra (LV; 2 processes) with 1024 samples, Table~\ref{tab:ablation_params} reports relative $L_2$ errors. While parameter-matched FNO improves over its base configuration, further widening or deepening yields diminishing returns and does not close the gap to COMPOL, indicating that the advantage primarily stems from explicit cross-process coupling rather than model size.

%

\paragraph{Improvement over Single-Process Simulations}
We further examine whether COMPOL’s latent aggregation mechanisms can enhance predictive performance for single-process simulations. Specifically, we evaluate COMPOL on the single-process Burgers’ equation using training sets of 512 and 1024 examples, comparing results against baseline methods (Table \ref{tab:ablation_single}). Results show that COMPOL significantly improves predictive accuracy, even in single-process scenarios, by effectively leveraging advanced latent aggregation techniques (attention-based and recurrent methods). These mechanisms enrich latent feature representations, implicitly regularize learning, adaptively select relevant features, and better capture underlying temporal dynamics. Consequently, COMPOL demonstrates improved optimization stability, superior generalization, and increased overall accuracy.

\paragraph{Predictive Spectrum Analysis.}
Energy spectra offer a physics-informed diagnostic of scale-wise fidelity in the Fourier domain. For each state variable, we compute the mean absolute error (MAE) between predicted and ground-truth spectral energy distributions (lower is better). As summarized in Table~\ref{tab:spectral-energy-mae}, \ours matches the ground-truth spectra most closely across both Belousov--Zhabotinsky (BZ; $u,v,w$) and Gray--Scott (GS; $u,v$) systems. Relative to FNO, \textsc{COMPOL-ATN} reduces spectral MAE on BZ by $\sim$5\% for $u$ and $\sim$28\% for $v$, while \textsc{COMPOL-RNN} yields the best $w$ with an $\sim$11\% drop. On GS, \textsc{COMPOL-RNN} improves over FNO by $\sim$2.5\% for $u$ and $\sim$7.7\% for $v$, whereas other baselines lag substantially (e.g., Transolver and DeepONet). These consistent gains indicate that coupling mechanisms in \ours help match the ground-truth spectral content across variables and systems.

\section{Conclusion}

We have introduced COMPOL, a novel coupled multi-physics neural operator learning framework that extends Fourier neural operators to effectively model interactions between multiple physical processes in complex systems. Our approach employs an attention-based feature aggregation mechanism to capture rich interdependencies among physical processes within the latent space, enabling explicit modeling of cross-process dynamics. Extensive experiments across diverse benchmarks, including reaction-diffusion systems, pattern-forming chemical reactions, multiphase geological flows, and thermo-hydro-mechanical processes, demonstrate that COMPOL achieves significant improvements in predictive accuracy compared to state-of-the-art methods. Ablation studies further confirm that these gains stem from the explicit cross-process coupling mechanism rather than increased model capacity. These results establish COMPOL as an effective and scalable framework for learning the complex dynamics of coupled multi-physics systems. 

While COMPOL demonstrates strong performance across the benchmarks studied, several limitations remain. First, all experiments use fixed structured grids, leaving COMPOL's ability to generalize across spatial resolutions unexplored; integrating resolution-invariant architectures is a natural extension. Second, the current evaluation focuses on single-step predictions from initial to final states; assessing stability under long-term autoregressive rollouts, where error accumulation is a known challenge, is an important direction for future work.


\section{Support of Open Science}
To support reproducibility, our complete codebase, including the COMPOL implementation, baseline comparisons, and all data generation scripts, is publicly available at \url{https://github.com/AriaQJ/COMPOL}.

\section*{Acknowledgment}
This work was supported in part by startup funding from the Department of Computer Science at Florida State University and the Cockrell School of Engineering at the University of Texas at Austin. Portions of this manuscript were revised with the assistance of Claude (Anthropic, \url{https://www.anthropic.com}) for language editing. and polishing.


\bibliographystyle{IEEEtran}
\bibliography{reference}


\end{document}